\documentclass[runningheads]{llncs}
\usepackage{graphicx}
\usepackage{amsmath} 
\usepackage{booktabs}
\usepackage{algorithm}
\usepackage{algorithmic}
\usepackage{verbatim}
\usepackage{multirow}
\usepackage{subfigure}
\title{Exploiting Dynamic and Fine-grained Semantic Scope for Extreme Multi-label Text Classification}
\titlerunning{Exploiting Dynamic and Fine-grained Semantic Scope for XMTC}
\author{Yuan Wang\inst{1,2} \and
Huiling Song\inst{1} \and
Peng Huo\inst{1} \and
Tao Xu\inst{1} \and
Jucheng Yang\inst{1} \and
Yarui Chen\inst{1} \and
Tingting Zhao\inst{1}}
\authorrunning{F. Author et al.}
\institute{College of Artificial Intelligence, Tianjin University of Science and Technology, Tianjin 300457, China \and
Population and Precision Health Care, Ltd., Tianjin 300000, China
}
\begin{document}
\maketitle              
\begin{abstract}
Extreme multi-label text classification (XMTC) refers to the problem of tagging a given text with the most relevant subset of labels from a large label set. A majority of labels only have a few training instances due to large label dimensionality in XMTC. To solve this data sparsity issue, most existing XMTC methods take advantage of fixed label clusters obtained in early stage to balance performance on tail labels and head labels. 
However, such label clusters provide static and coarse-grained semantic scope for every text, which ignores distinct characteristics of different texts and has difficulties modelling accurate semantics scope for texts with tail labels. In this paper, we propose a novel framework TReaderXML for XMTC, which adopts dynamic and fine-grained semantic scope from teacher knowledge for individual text to optimize text conditional prior category semantic ranges. TReaderXML dynamically obtains teacher knowledge for each text by similar texts and hierarchical label information in training sets to release the ability of distinctly fine-grained label-oriented semantic scope.
Then, TReaderXML benefits from a novel dual cooperative network that firstly learns features of a text and its corresponding label-oriented semantic scope by parallel Encoding Module and Reading Module, secondly embeds two parts by Interaction Module to regularize the text's representation by dynamic and fine-grained label-oriented semantic scope, and finally find target labels by Prediction Module. Experimental results on three XMTC benchmark datasets show that our method achieves new state-of-the-art results and especially performs well for severely imbalanced and sparse datasets.
\end{abstract}

\section{Introduction}
Recent years have witnessed remarkable progress in XMTC, with a variety of approaches presented in the literatures and applied in real-world scenarios, such as dynamic search advertising \cite{Prabhu:2014} and query recommendation \cite{Jain:2019}.

Different from classical multi-label problems, only a few are head labels with sufficient positive training data, and most labels are tail labels with few positive training data due to large label dimensionality \cite{Tagami:2017,Khandagale:2020,Chang:2020} in XMTC. This data sparsity issue leads to insufficient feature learning of tail labels, and hurts prediction performance on overwhelming tail label predictions. 

To solve this problem, most existing XMTC methods \cite{Jain:2016,Tagami:2017,Khandagale:2020,Prabhu:2014,Yashoteja:2018,Kush:2015,Jain:2016,Siblini:2018} take advantage of label clusters obtained in early stage to balance performance on tail labels and head labels. The main motivation is that the semantics of head labels is easy to be recognized in the semantic space due to sufficient training data, while the semantics of tail labels is vague. The precise semantics of tail labels can be learned from head labels that may appear in the same cluster. However, existing label clusters are all pre-defined global category patterns due to fixed features of labels. The static and coarse-grained semantic scope provided by such label clusters is not always consistent with dynamic real-world semantic scenarios, where content of different text has different semantic granularity. The previous model establishes structures hierarchy for the labels of a single field, and if the user is likely to be interested in overlapping topics in that and other fields, then when he enters a query into the search engine, he only gets keywords for a single domain due to the prepared label clusters.Thus, we consider developing dynamic semantic scope in the form of  fine-grained teacher knowledge to improve tail label predictions accuracy and alleviate the data sparsity issue.We introduce text relevance to increase exposure of tail labels and implement a dynamic label cluster structure to personalise relevant label subsets. In detail, for given instance, We can use the relevant labels of its neighbouring text to link more rare labels. We assume that if a text is related to a label, then the text is also related to its parent label. With the help of hierarchical label information, teacher knowledge is modeled to provide dynamic and fine-grained semantic scope to rich text semantics.

In summary, We propose a novel framework TReaderXML for XMTC containing a novel dual cooperative network based on multi-head self attention mechanism to embed both guidance knowledge and text into a shared semantic space for feature interaction, effectively improving the effect of teacher knowledge. The remainder of the paper is organized as follows. In Section 2, we review recent related work. Section 3 introduces TReaderXML. In Section 4, experimental results on three XMTC benchmark datasets are shown. Section 5 concludes this work.

\section{Related Work}
Many methods have been proposed for addressing the data sparsity issue of XMTC. They can be categorized into the following two types: 1) flat based label clusters \cite{Kush:2015,Tagami:2017}; 2) tree based label clusters \cite{Prabhu:2014,Jain:2016,Siblini:2018,Yashoteja:2018,You:2019,Khandagale:2020}. Tree based label clusters include loss function-based and structure-based.

In flat based label clusters, SLEEC \cite{Kush:2015} uses text features for clustering. A new text is projected in corresponding clusters, and labels of a new text are obtained by K-Nearest Neighbor to alleviate the data sparsity issue. Based on SLEEC, AnnexML \cite{Tagami:2017} uses label features for clustering based on graph embedding to improve the quality of clusters. In addition, in tree based label clusters, loss function-based method FastXML \cite{Prabhu:2014} learns an ensemble of trees which clusters the label space by optimizing a normalized Discounted Cumulative Gain (nDCG) loss function, and PfastreXML \cite{Jain:2016} replaces the nDCG loss in FastXML by its propensity scored variant which assigns higher rewards for tail label predictions. Furthermore, for structure-based methods in tree based label clusters, Parabel \cite{Yashoteja:2018} generates a label tree by recursively clustering labels into two balanced groups to address the data sparsity issue. However, the clustering depth of Parabel is deep, which leads to error cascade problems and affects tail label predictions. Bonsai \cite{Khandagale:2020} uses shallow and diverse probabilistic label trees (PLTs) by removing the balance constraint in the tree construction of Parabel, which improves tail label predictions. This tree structure-based label cluster optimization is also applied to AttentionXML \cite{You:2019}. AttentionXML optimizes the structure of PLTs to obtain shallow and wide clusters, which improves tail label predictions.

These label cluster methods provide static and coarse-grained semantic scope for every text. It is not always consistent with dynamic real-world semantic scenarios, and reduces the precision of prior knowledge.

\section{Methodology}
\subsection{Notation}
Given a training set $\{(x_{i}, y_{i})\}_{i=1}^{N}$ where $x_i$ is text input sequence, and $y_i\in\{0,1\}^{L}$ is the label of $x_i$ represented by $L$ dimensional multi-hot vectors. Each dimension in $y_i$ corresponds to a label where $y_{ij}=1$ when the j-th label $L_{y_{ij}}$ is associated with $x_i$. In this paper, we introduce teacher knowledge in a training set $\{(x_{i}, y_{i}, y_{i}^{\prime})\}_{i=1}^{N}$ where $y_{i}^{\prime}$ represents the text $x_i$'s corresponding teacher knowledge.

\subsection{TReaderXML}
\begin{figure}
\includegraphics[width=\textwidth]{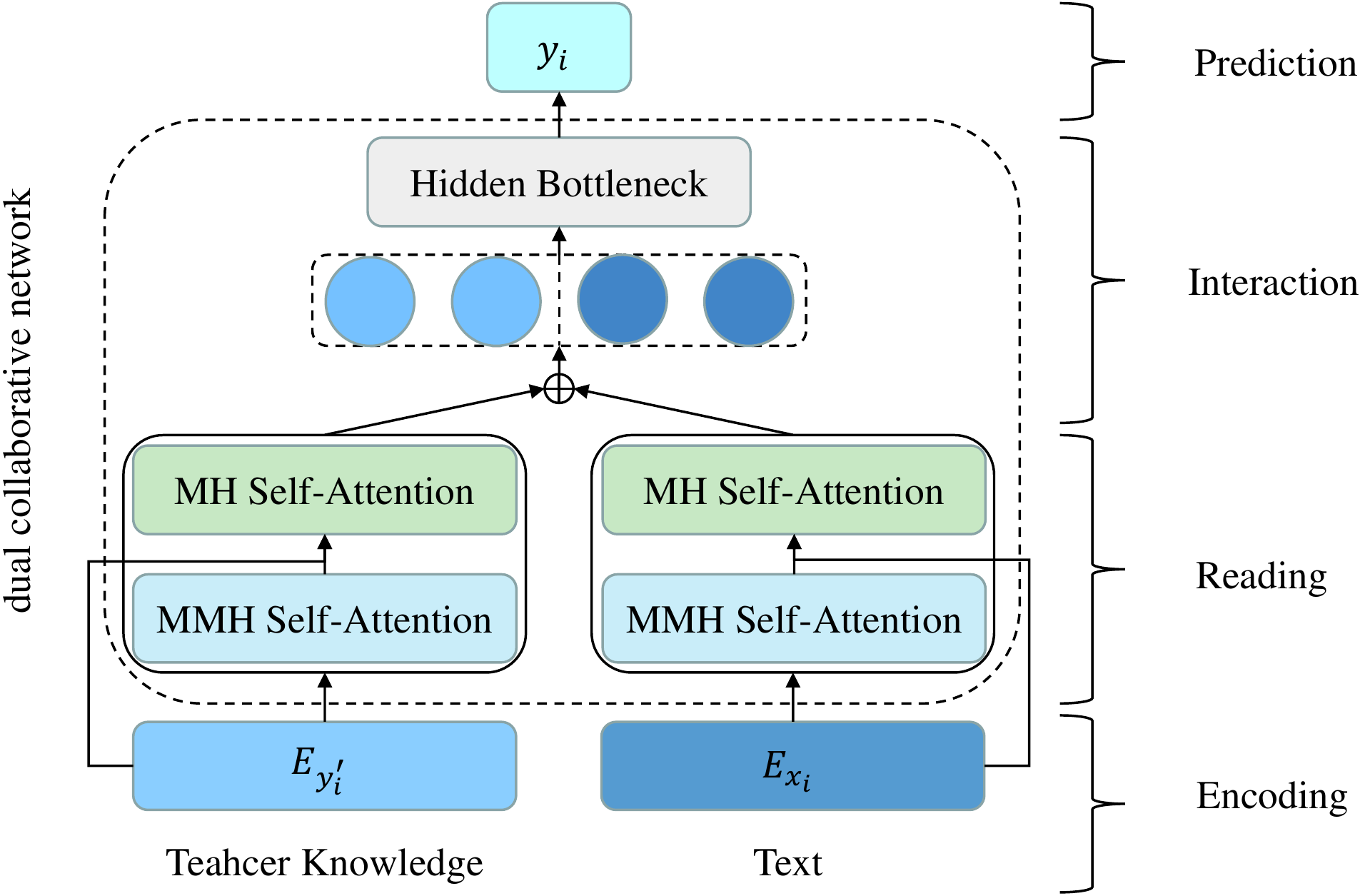}
\caption{An Overview of TReaderXML.} \label{Fig:2}
\end{figure}

TReaderXML adopts dynamic and fine-grained semantic scope from teacher knowledge for an individual text to optimize text prior category semantic ranges. Before teacher knowledge helps read text semantics, we need a powerful feature extraction to obtain high dimensional features of semantic scope from teacher knowledge and a text respectively, and embed both of them into a shared semantic space. Then the high dimensional semantics of scope with prior knowledge helps read high dimensional semantics of a text. Based on the above motivations, we design four layers: 1) Encoding, 2) Reading, 3) Iteraction and 4) Predicting. Furthermore, a dual cooperative network contains two layers of Reading and Iteraction, and Fig.~\ref{Fig:2} shows the framework of TReaderXML.
\subsubsection{Encoding}
In this part, we design a structure of representation to obtain fine-grained semantic scope extended by teacher knowledge matrix $E_{y_i'}$ and $E_{x_i}$. Given a training text ${x_{i}}$, 
its vectorization is shown as follows:
\begin{equation}
V_{x_{i}}=\frac{\sum_{c=1}^{{Len}({x_{i}})} \operatorname{Encode}(\mathrm{x}_{ic})}{{Len}(\mathrm{x_{i}})}.
\end{equation}
\vspace{-0.1cm}

Traverse ${x_{z}}$ from training set and validation set to find the nearest neighourhoods of ${x_{i}}$
by using consine similarity:
\begin{equation}
score_{cos}(V_{x_{i}},V_{x_{z}})=\frac{{(V_{x_{i}} \cdot V_{x_{i}})}}{\parallel V_{x_{i}}\parallel \parallel V_{x_{z}}\parallel}.
\end{equation}
and return top k nearest neighbourhoods $y_{i}^{nearest}$.To get the semantic scope, Gargiulo \cite{Gargiulo:2019} uses all of ancestor labels of text labels $y_i$ in a label tree to introduce hierarchical label information, effectively utilizing label semantic structural relation information. However, it leads to error cascade problems \cite{Khandagale:2020} due to the deep hierarchical structure. Furthermore, the semantics of deep hierarchical labels is often abstract, and it reduces the precision of prior knowledge. Inspired by these observations, we only use parent labels of child labels $y_i$ in a hierarchical label tree to introduce hierarchical label information. With the advantage of low error and high precision of hierarchical label information, teacher knowledge is modeled to provide dynamic and fine-grained semantic scope to help read text semantics.As shown in Algorithm 1, we firstly find the most relevant labels of $x_{i}^{nearest}$ and its non-empty parent labels. And then we put them into the label subset $SET^{nearest}$. Each label description information can be generated with widely used tricks in Parabel \cite{Yashoteja:2018}.
To keep an input sequence consistent with the semantic scope of teacher knowledge, we also initialize an embedding for an input sequence $x_i$, and the processing formula is shown as: $E_{x_i}=\text{Encode}(x_i)$.
\begin{algorithm}[htb]
\caption{Encoding generation of teacher knowledge}
\label{alg:TL}
\textbf{Input}: $y_{i}^{nearest}$\\
\textbf{Output}: $E_{y_i'}$
\begin{algorithmic}[1] 
\STATE initialize $SET^{nearest}$=\{\};
\FOR{$j=0$; $j<L$; $j++$ }
\IF {$y_{ij}^{nearest}==1$ and $Par(L_{y_{ij}}^{nearest})!=NULL$}
\STATE add $L_{y_{ij}}^{nearest}$ to $SET^{nearest}$
\STATE add $Par(L_{y_{ij}}^{nearest})$ to $SET^{nearest}$
\ENDIF 
\ENDFOR
\STATE \textbf{compute}
$E_{y_{i}^{\prime}}=\frac{\sum_{k=1}^{K} \operatorname{Encode}{(T_{k}^{nearest})}}{K}$\\
\STATE \textbf{return} $E_{y_i'}$
\end{algorithmic}
\end{algorithm}
\subsubsection{Reading}
In this part, we design a structure of Reading to obtain high dimensional features of semantic scope from a teacher knowledge and a text respectively and embed both of them into a shared semantic space for the preparation of feature interaction.This component in a dual cooperative network plays a key role including a mask multi-head self attention (MMHSA) layer, a multi-head self attention (MHSA) layer and residual network.

To obtain high dimensional features of semantic scope from teacher knowledge and a text respectively, we design the structure of Reading based on the self-attention mechanism \cite{Vaswani:2017}, which contains a MMHSA layer, a MHSA layer and a residual layer. MMHSA masks the future sequence information, and depends on existing sequence information to predict the next word in a sequence. We consider MMHSA as the first layer of Reading to capture more fine-grained semantic information due to the masking in MMHSA. Furthermore, MHSA makes each word contain other semantic information of words in a text input sequence, and we consider MHSA as the second layer of Reading to capture overall semantic information. The processing formula of masking in MMHSA is shown as follows:
\begin{align}
& d_{k}=d_{\text {model}} / / h, \\
& Q_{y_i'}=E_{y_i'}W_{y_i'}^{q},\text{ } Q_{x_i}=E_{x_i}W_{x_i}^{q}, \\
& K_{y_i'}=E_{y_i'}W_{y_i'}^{k},\text{ } K_{x_i}=E_{x_i}W_{x_i}^{k}, \\
& V_{y_i'}=E_{y_i'}W_{y_i'}^{v},\text{ } V_{x_i}=E_{x_i}W_{x_i}^{v}, \\
& Score_{y_i'}=\frac{Q_{y_i'} \cdot K_{y_i'}^{T}}{\sqrt{d_{k}}},\text{ } Score_{x_i}=\frac{Q_{x_i} \cdot K_{x_i}^{T}}{\sqrt{d_{k}}}, \\
& Score_{y_i'}=Mask(Score_{y_i'}, W_{y_i'}^{mask}), \\
& Score_{x_i}=Mask(Score_{x_i}, W_{x_i}^{mask}), \\
& H_i^{y_i'}=\operatorname{Softmax}(Score_{y_i'}) \cdot V_{y_i'}, \\
& H_i^{x_i}=\operatorname{Softmax}(Score_{x_i}) \cdot V_{x_i}.
\end{align}
where $d_{model}$ is dimension of embedding, and $h$ is the number of attention heads. $W_{y_i'}^q$, $W_{y_i'}^k$, $W_{y_i'}^v$, $W_{x_i}^q$, $W_{x_i}^k$, and $W_{x_i}^v$ are weight matrices of random initialization. $W_{y_i'}^{mask}$ and $W_{x_i}^{mask}$ are upper triangular matrices. For $Mask(A,B)$, positions where the value of $B$ is 0 are mapped into $A$, and the value of these positions are set to minus infinity in $A$. Then infinity values in $A$ will become 0 after Softmax, and masking has been achieved. The attention output $H_i^{y_i'}$ and $H_i^{x_i}$ learned by each head will be concatenated and transformed by multiplying a vector respectively. The output of MMHSA can be expressed by the formula given below:
\begin{align}
& MMHSA_{E_{y_i'}}=\operatorname{tanh}\left(\left\{H_{1}^{y_i'}, \ldots ; H_{h}^{y_i'}\right\} \cdot W_{y_i'}^{MH}\right), \\
& MMHSA_{E_{x_i}}=\operatorname{tanh}\left(\left\{H_{1}^{x_i}, \ldots ; H_{h}^{x_i}\right\} \cdot W_{x_i}^{MH}\right).
\end{align}

Compared with MMHSA, the processing formula of MHSA omits formulas (6) and (7).
To compensate for the loss of semantic information caused due to the masking techniques, we therefore introduce a residual network to enhance the robustness of the model and the expressiveness of the network:
\begin{align}
& E_{y_i'}^{residual}=E_{y_i'}+MMHSA(E_{y_i'}),\\
& E_{x_i}^{residual}=E_{x_i}+MMHSA(E_{x_i}).
\end{align}
The design of Reading simulates the process of reading a text. Firstly a teacher and a student respectively read verbatim to understand details of texts with MMHSA, and then read comprehensively to understand themes of texts with MHSA. Furthermore, the first layer of a dual cooperative network simulates the preparation of a teacher teaching a student to read. A teacher will prepare the key points of a text and a student will preview a text to achieve the best performance of reading. The first preparation work for the process of cooperation in a dual cooperative network has been achieved, and both semantic scope and text have been embedded into a shared semantic space.

\subsubsection{Iteraction}
The high-dimensional semantic scope generated by the interaction process between the teacher knowledge and the text provides a deeper understanding of the semantics of the text.

We assume that $O_{y_{i}'}$ represents the output of Reading from teacher knowledge, and $O_{x_{i}}$ represents the output of Reading from an input sequence. 
Firstly $O_{y_{i}'}$ and $O_{x_{i}}$ are concatenated, then prior knowledge helps read text semantics with a MMHSA layer like the first layer in Reading. The processing formula is shown as follows:
\begin{align}
& O_{concat}=[O_{y_{i}'} ; O_{x_{i}}],\\
& O_{Interaction}= O_{concat}\cdot W^{HB}.
\end{align}
where $W^{HB}$ represents the weight matrix of the hidden bottleneck layer. The bottleneck layer is properly constructed to significantly reduce the model size without degrading the network performance.
The design of Interaction containing a MMHSA layer simulates the process of a teacher teaching a student to read word by word. 
The second cooperation work in a dual cooperative network has been achieved, and the semantics of semantic scope and text has been enhanced in this network for better label prediction.

\subsubsection{Predicting}
Finally, a softmax layer is applied to predict final labels. The processing formula is shown as follows: 
\begin{align}
& Y = \text{Softmax}(O_{Interaction} \cdot W^{Output}).
\end{align}
where $W^{Output}$ is the output weight matrix of fully connected layer.

\subsubsection{Loss Function}
We measure the performance with multi-label one-versus-all loss based on max entropy principle, which are widely used in classification taskes. Specifically, for a predicted score vector $Y$ and a ground truth label vector $y_{i}$, the processing formula is shown as follows:
\begin{align}
Loss_i(Y, y_{i})=& -\sum_{j=1}^{L} y_{ij} \times \log \left((1+\exp (-Y_j))^{-1}\right)+ \nonumber\\
& (1-y_{ij}) \times \log \left(\frac{\exp (-Y_j)}{1+\exp (-Y_j)}\right) .
\end{align}

\section{Experiments}
\subsection{Datasets and Preprocessing}
\subsubsection{Datasets}

\begin{table}
\caption{Data statistics of three XMTC datasets.}\label{tab:1}
\resizebox{\textwidth}{!}{ 
\begin{tabular}{|c|c|c|c|c|}
\hline
Dataset       & Number of Train Points & Number of Test Points & Label Dimensionality & Avg. Labels per Point \\ \hline
AmazonCat-13K & 1,186,239              & 306,782               & 13,330               & 5.04                  \\ \hline
EURLex        & 15449                  & 3865                  & 3956                 & 5.30                  \\ \hline
RCV1          & 23,149                 & 781,265               & 103                  & 3.18                  \\ \hline
\end{tabular}
}
\end{table}

Three XMTC benchmark datasets, which have rich hierarchical information and label descriptionare, used for experiments in this paper, including AmazonCat-13K \cite{Bhatia:2016}, EURLex \footnote{\url{ http://manikvarma.org/downloads/XC/XMLRepository}} and RCV1 \cite{Lewis:2004}. Table~\ref{tab:1} shows the statistics of three datasets.

\subsubsection{Preprocessing Details}
For AmazonCat-13K, we truncate each input sequence after 300 words, and label description after 4 words in the same way as Parabel \cite{Yashoteja:2018}. Word embedding in AmazonCat-13K we use comes from AttentionXML \cite{You:2019}. For EURLex, we truncate each input sequence after 500 words, and each label description after 4 words. Word embedding in EURLex we use also comes from datasets. For RCV1, we truncate each input sequence after 250 words, and each label description after 16 words. Pre-trained Word2Vec \cite{Goldberg:2014} word embedding of 400 dimensions is used in RCV1.

The results of most these baseline methods are obtained from XMTC papers \cite{Liu:2017,Xiao:2019,You:2019}, and we have replicated unpublished results with original papers' codes. The word embedding training of RCV1 refers to methods \cite{rehurek_lrec:2010,Goldberg:2014}. The evaluation function implementation refers to the paper \cite{Huang:2019}. The framework of model training refers to the method \cite{Devlin:2018}. The experimental code on tail labels refers to AttentionXML \cite{You:2019}. The implementation of MHSA refers to the paper \cite{Zeng:2019}, and the number of attention heads $h$ in TReaderXML is set to 4. The initial learning rate for TReaderXML training is 0.0001. After the model converges, learning rate attenuation is used to further improve scores, and Adam \cite{Kingma:2014} is used for all deep learning model training. Our experimental configuration has a GPU of RTX 2080 Ti, and 128GB memory. When duplicating AnnexML \cite{Tagami:2017} on EURLex dataset, it cannot be duplicated due to memory problems.
\subsection{Baselines}
We compare our proposed TReaderXML to the most representative XMTC methods that address data sparsity issue including AnnexML \cite{Tagami:2017}, PfastreXML \cite{Jain:2016}, Parabel \cite{Yashoteja:2018}, FastText \cite{Joulin:2016}, Bonsai \cite{Khandagale:2020}, XML-CNN \cite{Liu:2017}, and AttentionXML \cite{You:2019}.
Table~\ref{tab:2} compares TReaderXML with baseline methods, and the results with stars are from XMTC papers \cite{Liu:2017,Xiao:2019,You:2019} directly.

The proposed TReaderXML outperforms all XMTC methods for most evaluation metrics, and for a few metrics it achieves results comparable to the current approaches.
Our method TReaderXML outperforms all XMTC methods, except for being slightly worse than LightXML (P@1) on AmazonCat-13K. Compared to leading extreme classifiers, TReaderXML can up to 0.16\% better in P@1 metric on RURLex. For the results of RCV1, TReaderXML has a substantial improvement at P@1. We consider that the precision of TReaderXML in the first predicting position is more accurate due to effective prior knowledge and TReaderXML remains close to existing XMTC methods in other evaluation metrics due to the small label dimensionality of RCV1.  

\subsection{Evaluation Metrics}
Classification accuracy is evaluated according to Precision at k (P@k), normalized Discounted Cumulative Gain at k (nDCG@k) and Propensity Scored Precision at k (PSP@k) like AttentionXML \cite{You:2019}.
refined
\subsection{Ablation Study}
We conduct an ablation study as shown in Table~\ref{tab:4} to discuss proposed novel structures of a dual cooperative network in TReaderXML. In detail, we explore the effectiveness of the teacher knowledge branch and the Reading part.
\subsubsection{Teacher Knowledge}
Config. ID 0, 1 shows the effectiveness of teacher knowledge. With dynamic and fine-grained semantic scope from teacher knowledge, Config. ID 1 has improved 5.2\% over Config. ID 0 without reading part.
\vspace{-0.4cm}
\begin{table}[]
\caption{Performance of TReaderXML and baseline methods over three datasets (The best results are highlighted in bold).}
\label{tab:2}
\resizebox{\textwidth}{!}{  
\begin{tabular}{|c|c|c|c|c|c|c|}
\hline
Datasets                        & Methods        & P@1              & P@3              & P@5              & nDCG@3           & nDCG@5           \\ \hline
\multirow{10}{*}{AmazonCat-13K} & AnnexML*       & 93.54\%          & 78.37\%          & 63.30\%          & 87.29\%          & 85.10\%          \\ \cline{2-7} 
                                & Parabel*       & 93.03\%          & 79.16\%          & 64.51\%          & 87.72\%          & 86.00\%          \\ \cline{2-7} 
                                & Bonsai*        & 92.98\%          & 79.13\%          & 64.46\%          & 87.68\%          & 85.92\%          \\ \cline{2-7} 
                                & PfastreXML*    & 91.75\%          & 77.97\%          & 63.68\%          & 86.48\%          & 84.96\%          \\ \cline{2-7} 
                                & XML-CNN*       & 93.26\%          & 77.06\%          & 61.40\%          & 86.20\%          & 83.43\%          \\ \cline{2-7} 
                                & AttentionXML*  & 95.92\%          & 82.41\%          & 67.31\%          & 91.17\%          & 89.48\%          \\ \cline{2-7} 
                                & X-Transformer* & 96.70\%          & 83.85\%          & 68.58\%          & -                & -                \\ \cline{2-7} 
                                & APLC-XLNet     & 94.56\%          & 79.82\%          & 64.61\%          & 88.74\%          & 86.66\%          \\ \cline{2-7} 
                                & LigntXML       & \textbf{96.77\%} & 84.02\%          & 68.70\%          & -                & -                \\ \cline{2-7} 
                                & TReaderXML     & 96.64\%          & \textbf{85.57\%} & \textbf{68.98\%} & \textbf{93.99\%} & \textbf{91.67\%} \\ \hline
\multirow{10}{*}{EURLex}        & AnnexML*       & 79.66\%          & 64.94\%          & 53.52\%          & 68.70\%          & 62.71\%          \\ \cline{2-7} 
                                & Parabel*       & 82.12\%          & 68.91\%          & 57.89\%          & 72.33\%          & 66.95\%          \\ \cline{2-7} 
                                & Bonsai*        & 82.30\%          & 69.55\%          & 58.35\%          & 72.97\%          & 67.48\%          \\ \cline{2-7} 
                                & PfastreXML*    & 73.13\%          & 60.16\%          & 50.54\%          & 63.51\%          & 58.71\%          \\ \cline{2-7} 
                                & XML-CNN*       & 68.01\%          & 54.03\%          & 43.93\%          & 57.44\%          & 51.83\%          \\ \cline{2-7} 
                                & AttentionXML*  & 87.12\%          & 73.99\%          & 61.92\%          & 77.44\%          & 71.53\%          \\ \cline{2-7} 
                                & X-Transformer* & 87.22\%          & 75.12\%          & 62.90\%          & -                & -                \\ \cline{2-7} 
                                & APLC-XLNet     & 87.72\%          & 74.56\%          & 62.28\%          & 77.90\%          & 71.75\%          \\ \cline{2-7} 
                                & LigntXML       & 87.63\%          & 75.89\%          & 63.36\%          & -                & -                \\ \cline{2-7} 
                                & TReaderXML     & \textbf{87.88\%} & \textbf{78.07\%} & \textbf{64.05\%} & \textbf{80.70\%} & \textbf{73.56\%} \\ \hline
\multirow{10}{*}{RCV1}          & AnnexML*       & 90.89\%          & 76.48\%          & 52.77\%          & -                & -                \\ \cline{2-7} 
                                & Parabel*       & 87.79\%          & 64.84\%          & 45.60\%          & 77.01\%          & 77.92\%          \\ \cline{2-7} 
                                & Bonsai*        & 85.23\%          & 65.12\%          & 45.89\%          & 76.55\%          & 77.59\%          \\ \cline{2-7} 
                                & PfastreXML*    & 68.82\%          & 60.76\%          & 43.28\%          & 69.40\%          & 71.24\%          \\ \cline{2-7} 
                                & XML-CNN*       & 93.63\%          & 73.90\%          & 52.16\%          & 85.24            & 86.69\%          \\ \cline{2-7} 
                                & AttentionXML*  & 96.41\%          & \textbf{80.91\%} & \textbf{56.38\%} & \textbf{91.88\%} & \textbf{92.70\%} \\ \cline{2-7} 
                                & X-Transformer* & -                & -                & -                & -                & -                \\ \cline{2-7} 
                                & APLC-XLNet     & 59.46\%          & 43.79\%          & 33.44\%          & -                & -                \\ \cline{2-7} 
                                & LigntXML       & 95.31\%          & 78.40\%          & 54.93\%          & -                & -                \\ \cline{2-7} 
                                & TReaderXML     & \textbf{97.50\%} & 78.74\%          & 54.67\%          & 90.29\%          & 90.94\%          \\ \hline
\end{tabular}
}
\end{table}
\vspace{-0.4cm}

\subsubsection{Reading}
Config. ID 2, 3, 4, 6 shows the plausibility of Reading structure. The structure of Config. ID 2 is similar to the effect of a person only reading word by word, and it cannot comprehensively understand themes of texts. The structure of Config. ID 3 is similar to the effect of a person only reading themes of texts, and it cannot carefully understand details of texts. The structure of Config. ID 4 is similar to the effect of a person reading themes of texts firstly then reading details of texts, and it is not always consistent with human reading habits. The structure of Config. ID 6 simulates the process of human reading, reading word by word to understand details of texts and reading comprehensively to understand themes of texts. It is feasible to simulate human reading with the Reading structure. Config. ID 5, 6 shows the effectiveness of residual layer. Config. ID 6 has improved 0.44\% over Config. ID 5 with residual part.

\vspace{-0.4cm}
\begin{table}
\caption{Ablation study of TReaderXML on AmazonCat-13K (The best results are highlighted in bold).}
\label{tab:4}
\resizebox{\textwidth}{!}{
\begin{tabular}{|c|c|c|c|c|c|c|c|}
\hline
Config. ID & teacher knowledge & Reading      & P@1              & P@3              & P@5              & nDCG@3           & nDCG@5           \\ \hline
0          & -                 & -            & 88.73\%          & 69.12\%          & 53.94\%          & 78.78\%          & 75.42\%          \\ \hline
1          & True              & -            & 93.93\%          & 77.70\%          & 58.53\%          & 87.15\%          & 81.79\%          \\ \hline
2          & True              & MMHSA+R      & 95.49\%          & 83.45\%          & 66.08\%          & 92.00\%          & 88.81\%          \\ \hline
3          & True              & MHSA+R       & 95.52\%          & 83.46\%          & 66.04\%          & 92.03\%          & 88.81\%          \\ \hline
4          & True              & MHSA+R+MMHSA & 96.49\%          & 85.44\%          & 68.84\%          & 93.86\%          & 91.52\%          \\ \hline
5          & True              & MMHSA+MHSA   & 96.20\%          & 84.89\%          & 68.04\%          & 93.33\%          & 90.72\%          \\ \hline
6          & True              & MMHSA+R+MHSA & \textbf{96.64\%} & \textbf{85.57\%} & \textbf{68.98\%} & \textbf{93.99\%} & \textbf{91.67\%} \\ \hline
\end{tabular}
}
\end{table}
\vspace{-0.4cm}

\subsection{Performance on tail labels}
To evaluate performance of TReaderXML on tail labels, we discuss experiment results of tail labels on AmazonCat-13K dataset which has the most tail labels. From Table~\ref{tab4}, we see that TReaderXML achieves SOTA effects at  PSP@5, except for being slightly worse than PfastreXML~\cite{Jain:2016} at PSP@1 and PSP@3. PfastreXML replaces the nDCG loss in FastXML~\cite{Prabhu:2014} by its propensity scored variant which is unbiased and assigns higher rewards for the tail label predictions. However, it leads to a loss in prediction accuracy.
\vspace{-0.4cm}
\begin{table}
\centering
\caption{Performance on tail labels in AmazonCat-13K (The best results are highlighted in bold).}\label{tab4}
\begin{tabular}{|c|c|c|c|}
\hline
Methods        & PSP@1            & PSP@3            & PSP@5            \\ \hline
AnnexML*       & 49.04\%          & 61.13\%          & 69.64\%          \\ \hline
Parabel*       & 50.93\%          & 64.00\%          & 72.08\%          \\ \hline
Bonsai*        & 51.30\%          & 64.60\%          & 72.48\%          \\ \hline
PfastreXML*    & \textbf{69.52\%} & \textbf{73.22\%} & 75.48\%          \\ \hline
XML-CNN*       & 52.42\%          & 62.83\%          & 67.10\%          \\ \hline
AttentionXML*  & 53.76\%          & 68.72\%          & 76.38\%          \\ \hline
X-Transformer* & -                & -                & -                \\ \hline
APLC-XLNet     & 52.22\%          & 65.08\%          & 71.40\%          \\ \hline
LigntXML       & \textbf{-}       & -                & -                \\ \hline
TReaderXML     & 57.15\%          & 71.64\%          & \textbf{77.27\%} \\ \hline
\end{tabular}

\end{table}
\vspace{-0.4cm}

\section{Conclusions}
In this work, our method TReaderXML define semantic scope from teacher knowledge, which inherits the strength of hierarchical label information and meanwhile improves dynamic high level category information as semantic supplements and constraints. The proposed dual cooperative network learned semantic information in the way of people reading. Moreover, teacher knowledge can flexibly incorporate prior label information like semantic structures or descriptions.
%
%
%
\bibliographystyle{splncs04}
\bibliography{mybibliography}
%




\end{document}